\documentclass[a4paper,twoside]{article}

\usepackage{epsfig}
\usepackage{subcaption}
\usepackage{calc}
\usepackage{amssymb}
\usepackage{amstext}
\usepackage{amsmath}
\usepackage{amsthm}
\usepackage{multicol}
\usepackage{pslatex}
\usepackage{apalike}
\usepackage{algorithm2e}
\usepackage[bottom]{footmisc}
\usepackage{amsfonts}
\usepackage{bm}
\usepackage{float}
\usepackage{graphicx}
\usepackage{array}
\usepackage{multirow}
\usepackage{amsbsy}
\usepackage{url}
\usepackage{colortbl}
\usepackage{hyperref}
\usepackage[table]{xcolor}

\usepackage{SCITEPRESS}     

\begin{document}

\title{On Spectrogram Analysis in a Multiple Classifier Fusion Framework for Power Grid Classification Using Electric Network Frequency}

\author{\authorname{Georgios Tzolopoulos\orcidAuthor{0009-0000-0405-889X}, Christos Korgialas\orcidAuthor{0000-0001-5475-0518} and Constantine Kotropoulos\orcidAuthor{0000-0001-9939-7930}}
\affiliation{Department of Informatics, Aristotle University of Thessaloniki, Thessaloniki 54124, Greece}
\email{\{gtzolopo, ckorgial, costas\}@csd.auth.gr}}


\abstract{The Electric Network Frequency (ENF) serves as a unique signature inherent to power distribution systems. Here, a novel approach for power grid classification is developed, leveraging ENF. Spectrograms are generated from audio and power recordings across different grids, revealing distinctive ENF patterns that aid in grid classification through a fusion of classifiers.  Four traditional machine learning classifiers plus a Convolutional Neural Network (CNN), optimized using Neural Architecture Search, are developed for One-vs-All classification.
This process generates numerous predictions per sample, which are then compiled and used to train a shallow multi-label neural network specifically designed to model the fusion process, ultimately leading to the conclusive class prediction for each sample. Experimental findings reveal that both validation and testing accuracy outperform those of current state-of-the-art classifiers, underlining the effectiveness and robustness of the proposed methodology.}

\onecolumn \maketitle \normalsize \setcounter{footnote}{0} \vfill

\section{\uppercase{Introduction}}
\label{sec:introduction}

The Electric Network Frequency (ENF) \cite{grigoras2005digital} serves as a ``fingerprint", potentially embedded in multimedia content, such as audio recordings, that are captured in proximity to the power mains \cite{cooper2009automated}. ENF fluctuates instantaneously around its nominal value of 60 Hz in the United States (US)/Canada or 50 Hz in the rest of the world. These small fluctuations in frequency hold great importance, providing invaluable insights into forensic applications \cite{grigoras2007applications}, \cite{ngharamike2023enf}. Such applications extend to device identification \cite{hajj2016exploiting}, \cite{bykhovsky2020recording}, \cite{ngharamike2023exploiting}, and verifying the timestamp of multimedia recordings \cite{hua2014dynamic}, \cite{garg2013seeing}, \cite{vatansever2022enf}.

A notable application of ENF is the ability to pinpoint the location of a recording at both inter-grid and intra-grid localization levels. Inter-grid localization capitalizes on the distinctive ENF signatures of different power grids, facilitating the determination of a recording's broader geographical region or grid of origin. Meanwhile, intra-grid localization focuses on the finer distinctions within a single power grid. Despite the inherently high similarity of ENF variations recorded concurrently at different locations within the same grid, discernible differences have been observed, rooted in city-specific power consumption changes and the time lags required for load-related variations to disseminate across the grid \cite{garg2013geo}, \cite{elmesalawy2014new}. Such discrepancies can also emerge from systemic disruptions like power line switching or generator disconnections. For instance, a localized load change might influence the ENF specifically in its vicinity, while a substantial system change, like a generator disconnection, has ramifications for the entire grid. This shift, intriguingly, propagates across the Eastern US grid at a staggering rate of roughly 500 miles per second \cite{tsai2007frequency}. To classify audio recordings captured in different power grids globally, the complexities of inter-grid characteristics have prompted experts to develop various strategies. Notably, events such as the 2016 Signal Processing Cup \cite{wu2016location} have showcased these methods, advancing ENF-based forensics and reinforcing the authenticity of multimedia recordings.

Here, inter-grid classification is tackled from the perspective of the fusion of multiple machine learning classifiers, including Logistic Regression, Naive Bayes (NB), Random Forest (RF), and Multilayer Perceptron (MLP), with an optimized Convolutional Neural Network (CNN) using Neural Architecture Search (NAS). To delve deeper, these classifiers are trained on the spectrograms derived from audio and power recordings captured in various grids. The validation accuracy achieved through a One-vs-All classification signifies the effectiveness of the proposed framework against the state-of-the-art methods for power grid classification. In a nutshell, by leveraging the strengths of multiple classifiers, the fusion model provides robustness against overfitting and results in improved generalization to unseen data, further enhancing the reliability of the proposed model.

The main contributions of the paper are as follows:
\begin{itemize}
    \item A fusion model is developed that combines five machine learning classifiers, including an optimized CNN by means of NAS.
    \item Data augmentation is applied to the audio and power recordings, which are then transformed into spectrograms, focusing on the nominal frequency of 50 or 60 Hz. 
    \item One-vs-All classification is utilized. Testing accuracy was calculated to evaluate the effectiveness of the proposed framework.
\end{itemize}

Related work is surveyed in Section~\ref{sec:related}, while in Section~\ref{sec:methodology} the proposed methodology is analyzed. In  Section~\ref{sec:results}, experimental results are presented and discussed. In Section~\ref{sec:conclusion}, conclusions and insights to future work are offered. 

\begin{figure*}[t]
 \centering
 {\epsfig{file = 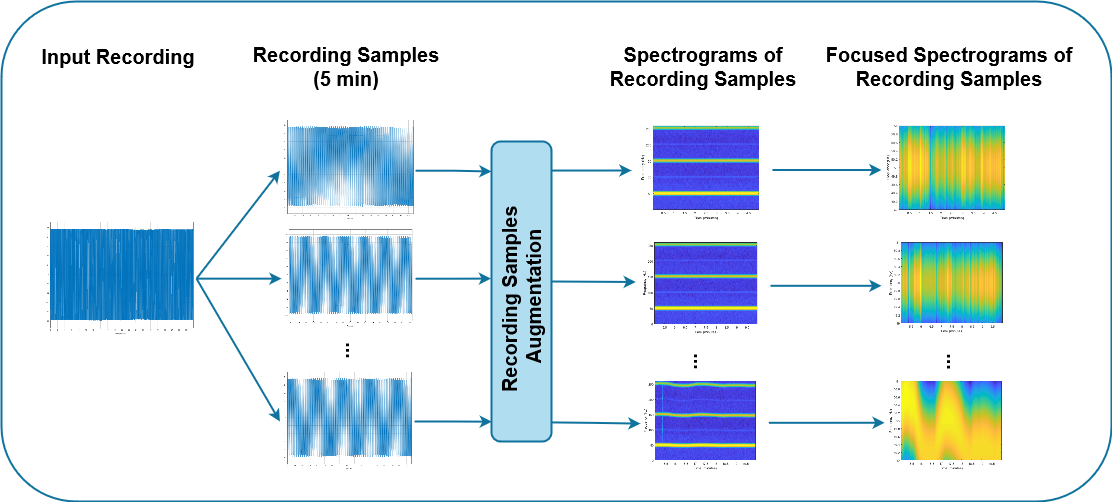, width = 14cm}}
  \caption{\centering Flowchart illustrating the preprocessing steps of the audio and power recordings.}
\label{fig:data_prepro}
\end{figure*}

\section{\uppercase{Related Work}}
\label{sec:related}

\subsection{Power Grid Location Estimation}

The fluctuating ENF is a potential tool for geolocation and power grid identification, leveraging its distinct imprints on multimedia recordings. Building upon the principles of ENF signal applications, \cite{hajj2013enf} enhanced grid identification with advanced machine learning techniques and an in-depth analysis of ENF variations. Subsequently, \cite{hajj2015enf} developed a multiclass machine learning model that leveraged statistical ENF variations to accurately determine the power grid locations of multimedia recordings, even in the absence of simultaneous power reference. Experiments conducted in \cite{garg2013geo} demonstrated that analyzing ENF fluctuation similarities, which correlate with geographic distance, can estimate multimedia recording locations with a high accuracy. In \cite{garg2021feasibility}, the potential of embedded ENF traces in multimedia recordings was evaluated to determine a recording's specific location within a power grid, showcasing that the correlation of high-pass filtered ENF signals decreases with greater geographic distance, thereby enabling the creation of trilateration-based localization techniques. Machine learning algorithms were utilized in \cite{vsaric2016improving}, particularly the RF, to classify ENF signals from various power grids, enhancing detection accuracy by introducing signal features. In \cite{sarkar2019application}, a location-stamp authentication method was introduced, employing ENF sequences from digital recordings to verify the specific location of recordings, substantiated by applying a multiclass Support Vector Machine (SVM) classification model. Distribution-level ENF data from the FNET/GridEye system were leveraged in \cite{yao2017source}, introducing a hybrid method combining wavelet-based signature extraction with neural network learning to trace the location origins of ENF signals accurately. In \cite{kim2020location}, an approach to pinpoint a multimedia file's playback location was presented by analyzing ENF signals from online streaming videos using a secondary interpolation, which enhances the resolution of ENF signals by applying quadratic interpolation to the results of a Short-Time Fourier Transform (STFT) and Autoregressive Integrated Moving Average (ARIMA) modeling bypassing the need for an additional interpolation step.

\subsection{Ensemble Learning for Audio Spectrogram Classification}

Ensemble learning, by combining multiple machine learning models, has shown great potential in audio spectrogram classification \cite{mienye2022survey}. In \cite{jiang2019acoustic}, 16 ensemble methodologies were employed to analyze audio recordings, with a particular focus on various spectrogram decomposition techniques. The accuracy of acoustic scene classification was significantly enhanced by combining CNNs and ensemble classifiers using late fusion, as demonstrated in \cite{alamir2021novel}, surpassing the performance of individual models. In \cite{le2019using}, machine learning methodologies and ensemble classification techniques were applied to differentiate various types of baby cries from spectrogram images, achieving high accuracy. In \cite{nanni2020ensemble}, an ensemble method was created for automated audio classification by fusing different features from audio recordings, improving accuracy over existing approaches, and marking a significant advancement in CNN-based animal audio classification. The effectiveness of the self-paced ensemble learning scheme, where models iteratively learn from each other, was significantly demonstrated in \cite{ristea2021self}, outperforming baseline ensemble models in three audio tasks.

\section{\uppercase{Methodology}}\label{sec:methodology}

In this Section, the composition of the dataset, as well as the preprocessing steps applied, are outlined. Moreover, the proposed fusion framework is described.

\subsection{Data Description and Preprocessing}\label{sec:data_desc}

Here, the dataset from the 2016 Signal Processing (SP) Cup \cite{data2016} is used. The dataset comprises recordings from nine distinct power grids, each labeled from $\bm{A}$ to $\bm{I}$. Recordings from grids $\bm{A}$, $\bm{C}$, and $\bm{I}$ include a 60 Hz ENF signal, while the remaining grids feature a 50 Hz nominal ENF. Moreover, audio recordings are included from a variety of settings, and power recordings are obtained through a special circuit, with durations varying from 30 to 60 minutes. The power recordings contain inherently stronger ENF traces, whereas audio recordings exhibit a higher degree of noise, rendering ENF utilization a more challenging endeavor. For testing purposes, 100 additional 10-minute long recordings, comprising 40 audio and 60 power recordings, are provided. The audio recordings are obtained by placing microphones near power devices to capture their characteristic hum. Some of these recordings belong to grids not included in the original nine ones and are thus to be classified as ``None” ($\bm{N}$).

Figure~\ref{fig:data_prepro} summarizes the preprocessing steps applied to the audio and power recordings from the power grids labeled $\bm{A}$ to $\bm{I}$. The figure depicts the transformation of raw audio data into a form suitable for the detailed analysis of ENF signals. Each recording is initially segmented into 5-minute samples, providing a uniform length for all samples. These samples are then augmented with white noise, specifically around the frequency band centered on 50 to 60 Hz with a $\pm$ 1 Hz tolerance. Subsequently, the augmented audio samples are transformed into spectrograms, with the yellow lines denoting the presence of the ENF signals at either 50 or 60 Hz, as well as their higher harmonics. The final preprocessing step involves focusing the spectrogram on the nominal value of ENF (i.e., 50 or 60 Hz), depending on the grid in question. The focused spectrograms will be utilized as inputs to the five classifiers in the fusion framework during both training and testing phases (see Section~\ref{subsec:fusion}).

\begin{figure*}[!ht]
  \centering
   {\epsfig{file = 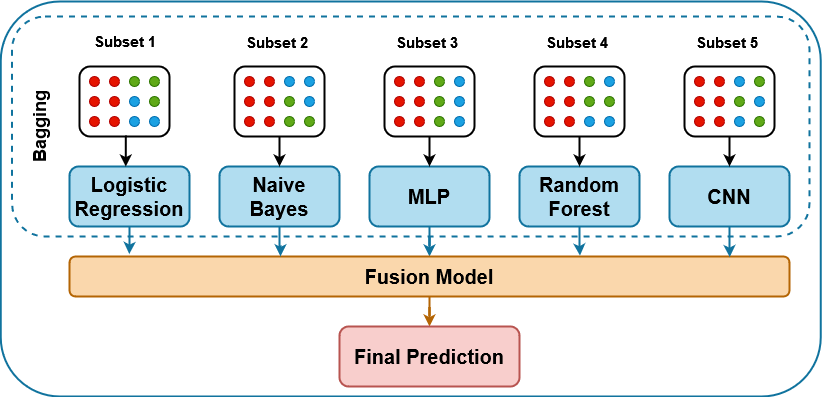, width = 14cm}}
  \caption{Flowchart of the proposed fusion framework for power grid classification.}
  \label{fig:fusion_framework}
 \end{figure*}

 \begin{table*}[t]
\centering
\caption{Architecture of the optimized CNN.}
\footnotesize 
\setlength{\tabcolsep}{45pt} 
\begin{tabular}{|c|c|c|}
  \hline
  & \textbf{Layer} & \textbf{Output} \\
  \hline
  \multirow{2}{*}{Layer 1} & Conv2D & $72,540 \times 32$ \\
  \cline{2-3}
  & MaxPool & $18,135 \times 32$ \\
  \hline
  \multirow{2}{*}{Layer 2} & Conv2D & $16,907 \times 64$ \\
  \cline{2-3}
  & MaxPool & $4,074 \times 64$ \\ 
  \hline 
  \multirow{2}{*}{Layer 3} & Conv2D & $3,467 \times 128$ \\
  \cline{2-3}
  & MaxPool & $864 \times 128$ \\
  \hline
  \multirow{3}{*}{Layer 4} & Flatten & $110,592$ \\ 
  \cline{2-3}
  & Dense & $101$ \\ 
  \cline{2-3}
  & Dropout & $101$ \\ 
  \hline
  Layer 5 & Dense & $1$ \\
  \hline 
\end{tabular}
\label{tab:cnn_arch}
\end{table*}

\subsection{Overview of Classifiers}\label{subsec:classifiers}

Here, the five classifiers integrated into the fusion model are described. A Logistic Regression model with an $\ell_2$ penalty and a regularization constant of 1.0 is chosen for its effectiveness in binary classification. A Naive Bayes classifier is employed, adjusted with a smoothing factor of $10^{-9}$ to improve performance on sparse data. An MLP featuring two hidden layers, the first with 100 neurons and the second with 50 neurons, is included. An RF classifier is incorporated, consisting of 100 trees allowed to be fully developed for data division, with feature selection conducted automatically. Most of the parameters utilized in the classifiers are sourced from the $\texttt{scikit-learn}$ library \cite{pedregosa2011scikit}. 

\begin{figure*}[!ht]
  \centering
   {\epsfig{file = 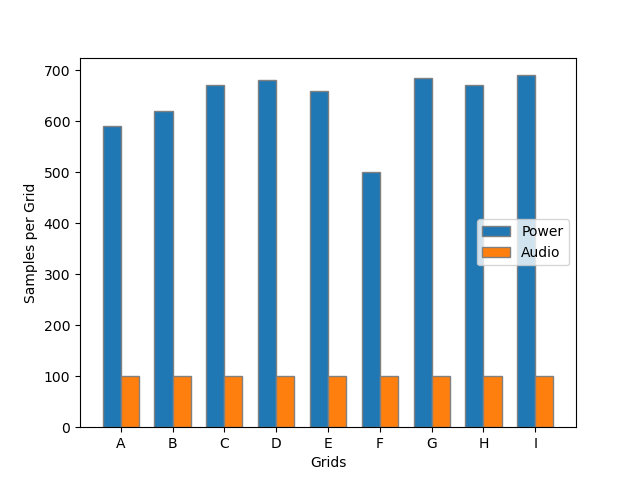, width = 12cm}}
  \caption{Number of audio and power recording frames in each grid.}
  \label{fig:samples}
\end{figure*}

In Table~\ref{tab:cnn_arch}, the optimized CNN architecture is presented, employing the NAS approach (see Section~\ref{subsec:training_testing}). The CNN's architecture progresses through sequential layers, starting with two-dimensional convolutions and max-pooling operations, which gradually reduce the spatial dimensions of the input spectrograms while increasing their depth to encapsulate more complex features. As the architecture advances, these refined two-dimensional feature maps are transformed into a one-dimensional vector through a flattening process. This vector is then processed by successive dense layers designed to interpret the features abstracted from the ENF spectrograms. A dropout layer is included to mitigate overfitting by randomly omitting a proportion of the input units during the training phase. The output layer, a single dense unit, is pivotal for the implementation of the `One-vs-All' classification strategy, allowing the model to predict the probability that a given ENF spectrogram belongs to one of nine classes $(\bm{A}$ to $\bm{I})$ by comparing it against all others, thus enabling the determination of the most likely class for each instance.

\subsection{Fusion Framework}\label{subsec:fusion}
 
The proposed framework is detailed (see Figure~\ref{fig:fusion_framework}), encompassing the data-splitting process and the fusion model description. 

The recordings fall into two independent categories: audio and power. Furthermore, grids with ENF at 50 Hz are distinguishable from those with ENF at 60 Hz. Consequently, the entire dataset is divided into four distinct sub-datasets: \texttt{audio50}, \texttt{audio60}, \texttt{power50}, and \texttt{power60}.

While this categorization is known during the training phase through the provided data description, testing requires developing methods to identify each sample's category. The distinction between audio and power can be perceived by human hearing, given that audio recordings exhibit a significantly lower signal-to-noise ratio (SNR) than power recordings. This characteristic can be leveraged to automate audio/power identification. Recordings containing ENF at 50 Hz exhibit higher frequency content in the bands near 50 Hz and in their harmonics. The same applies to recordings containing ENF at 60 Hz. Thus, a method was devised to compare the magnitude of the Fourier Transform at the first harmonic of the recordings for both nominal frequencies, enabling the determination of the ENF of a recording.

The dataset is divided into four independent subsets in the data-splitting process. This methodology applies to all classes, regardless of the number of resulting classes. The classification framework is structured around classes $G^{\mathrm{ENF}}_{\mathrm{REC}} = \{C_1, C_2, \dots, C_n\}$ \footnote{For ease of notation, the term $G^{\mathrm{ENF}}_{\mathrm{REC}}$ will be referred to as $G$.}, where $n=3$ for class from grids with ENF in 60Hz and $n=6$, otherwise. Let $C_i$ consist of data samples $\textbf{x}_{i,j}$, where $i$ indicates the class index and $j$ is the sample index within that class. These samples are characterized by having the same nominal ENF and recording type (REC).

For illustration, let us consider the \texttt{audio60} sub-dataset, which contains audio recordings from grids $\bm{A}$, $\bm{C}$, and $\bm{I}$. The classification challenge then narrows down to $G^{60}_{\mathrm{audio}}$. Consequently, the training dataset is defined as $\mathbf{X} = \{\mathbf{x}_{i,j} \; | \; \mathbf{x}_{i,j} \in C_i, \forall \; i \in \{1, 2, \dots, n \}  \}$ for $n=3$, with each $\mathbf{x}_{i,j}$ representing a data sample in class $C_{i}$, for $n=3$. The corresponding label set is $\mathbf{Y} = \{{y}_{i,j} \; | \; {y}_{i,j} \in C_i, \forall \; i \in \{1, 2, \dots, n \}  \}$, where ${y}_{i,j}$ is the label associated with sample $\mathbf{x}_{i,j}$.

For audio recordings, samples are uniformly distributed across classes (see Figure~\ref{fig:samples}). However, this uniformity does not hold for power recordings. Such uneven distribution could potentially lead to overfitting in favor of a class with more samples while training a multi-label model. To mitigate this, a One-vs-All strategy with $|G|$ models is devised, where $|G|$ stands for the cardinality, indicating the total number of distinct classes in $G$. Each model, denoted as $M_i$, is designed to separate samples of class $C_i$ from samples of classes $\{C_j \; | \; i \neq j \}$. During the training of $M_i$,  all training samples of class $C_i$ are utilized, along with an equal number of samples from each class $C_j$,  $i \neq j$, maintaining collinearity with the number of samples in $C_i$. This results in balanced training sets for each model, posing a binary classification problem for each.

\begin{table*}[!ht]
\centering
\caption{Comparison of average validation accuracies (\%) for five classifiers and the fusion model.}
{
\setlength{\tabcolsep}{7.5pt}
\begin{tabular}{|c|c|c|c|c|c|c|c|c|c|c|}
\hline
\textbf{Type} & \textbf{Classifier} & $\bm{A}$ & $\bm{B}$ & $\bm{C}$ & $\bm{D}$ & $\bm{E}$ & $\bm{F}$ & $\bm{G}$ & $\bm{H}$ & $\bm{I}$ \\
\hline
\multirow{6}{*}{\rotatebox{90}{Audio}} 
& Naive Bayes & 72.4 & 67.5 & 78.8 & 75 & 75 & 95 & 80 & 52.5 & 87.5\\
\cline{2-11}
& Logistic Regression & \cellcolor{gray!25} 100 & 95 & 85 & 95.5 & \cellcolor{gray!25} 100 & 95 & 92.5 & 82.5 & \cellcolor{gray!25} 100\\
\cline{2-11}
& RandomForest & 95 & 92.5 & 87.5 & 97.5 & \cellcolor{gray!25} 100 & \cellcolor{gray!25} 100 & 97.5 & 85 & 97.5\\
\cline{2-11}
& MLP & 97.5 & 85 & 85 & 47.5 & 95 & 95 & 85 & 87.5 & \cellcolor{gray!25} 100\\
\cline{2-11}
& CNN & 98.3 & 95 & 94.7 & \cellcolor{gray!25} 100 & 99.3 &  \cellcolor{gray!25} 100 & \cellcolor{gray!25} 99.1 & 92.8 & 99.3\\
\cline{2-11}
& \textbf{Fusion Model} & \cellcolor{gray!25} 100 & \cellcolor{gray!25} 97.4 & \cellcolor{gray!25} 98.1 & \cellcolor{gray!25} 100 & \cellcolor{gray!25} 100 & \cellcolor{gray!25} 100 & 98.2 & \cellcolor{gray!25} 95.6 & \cellcolor{gray!25} 100\\
\hline
\multirow{6}{*}{\rotatebox{90}{Power}} 
& Naive Bayes & 79.7 & 81.9 & 85.63 & 89.9 & 84.8 & 88 & 90.63 & 61.9 & 79.6\\
\cline{2-11}
& Logistic Regression & 98.8 & 99.25 & 97.4 & 97.3 & 97 & 98.5 & 91.6 & 98.5 & \cellcolor{gray!25} 100\\
\cline{2-11}
& RandomForest & \cellcolor{gray!25} 100 & \cellcolor{gray!25} 100 & 99.3 & \cellcolor{gray!25} 100 & \cellcolor{gray!25} 100 & \cellcolor{gray!25} 100 & 97.8 & 99.6 & 99.6\\
\cline{2-11}
& MLP & \cellcolor{gray!25} 100 & \cellcolor{gray!25} 100 & 98.9 & 98.9 & 95 & \cellcolor{gray!25} 100 & 98.2 & 98.9 & \cellcolor{gray!25} 100\\
\cline{2-11}
& CNN & 99 & \cellcolor{gray!25} 100 & \cellcolor{gray!25} 100 & \cellcolor{gray!25} 100 & 99.3 & \cellcolor{gray!25} 100 & 98.1 & \cellcolor{gray!25} 100 & \cellcolor{gray!25} 100\\
\cline{2-11}
& \textbf{Fusion Model} & \cellcolor{gray!25} 100 & \cellcolor{gray!25} 100 & 99.7 & \cellcolor{gray!25} 100 & \cellcolor{gray!25} 100 & \cellcolor{gray!25} 100 & \cellcolor{gray!25} 98.4 & \cellcolor{gray!25} 100 & \cellcolor{gray!25} 100\\
\hline
\end{tabular}}
\hfill{}
\label{tab:validation_acc}
\end{table*}

\begin{table*}[!ht]
\centering
\caption{Fusion framework accuracy (\%) in the testing set.}
{
\setlength{\tabcolsep}{9.8pt}
\begin{tabular}{|c|c|c|c|c|c|c|c|c|c|c|c|}
\hline
\textbf{Type} & $\bm{A}$ & $\bm{B}$ & $\bm{C}$ & $\bm{D}$ & $\bm{E}$ & $\bm{F}$ & $\bm{G}$ & $\bm{H}$ & $\bm{I}$ & $\bm{N}$ & \textbf{Overall}\\
\cline{3-9}
\hline
Audio & 100 & 100 & 100 & 100 & 100 & 100 & 100 & 75 & 100\ & 25\ & 90\\
\hline
Power & 100 & 100 & 100 & 100 & 100 & 100 & 100 & 100 & 100\ & 100\ & 100\\
\hline
All & 100 & 100 & 100 & 100 & 100 & 100 & 100 & 90 & 100\ & 70\ & 96\\
\hline
\end{tabular}}
\hfill{}
\label{tab:testing_acc}
\end{table*}

No individual classifier among those described in Section~\ref{subsec:classifiers} yields adequate accuracy, as seen in Table~\ref{tab:validation_acc}.  The same One-vs-All strategy is applied to each classifier. To introduce diversity, the bagging technique is employed \cite{breiman1996bagging}, utilizing different data subsets corresponding to classes $C_j$, where $i \neq j$, for training $M_i$ for every classifier. Figure~\ref{fig:fusion_framework} illustrates that each classifier is trained using a separate data subset. This approach utilizes all available data, enhancing the generality of the final model.

For the final class prediction of a sample, a fusion of decisions (depicted by the orange box in Figure~\ref{fig:fusion_framework}) from the individual models is necessary.
This process combines the strengths of all classifiers, contributing to creating a robust final model that summarizes the knowledge encoded in the models. 
Each sample under analysis generates a substantial number of $5 \times |G|$ predictions.  This number arises from combining five distinct classifiers, each contributing  $|G|$ predictions. 

The framework also addresses the challenge posed by different durations of training and testing samples. While training samples are of 5-minute length, testing samples are twice as long, requiring them to be split into two separate 5-minute segments. Consequently, this splitting leads to a total of $2 \times 5 \times |G|$ predictions for each testing sample, encompassing predictions for each segment by each classifier across all models. 
These estimations describe the assessment of a sample from various perspectives, as indicated by different classifiers. 

Fusing all the predictions is necessary to distill meaningful insights from this diverse information. The fusion process extracts the final prediction for the grid of the recording.
Thus, all the predictions for one sample are then compiled into vectors of size $10 \times |G|$, which are instrumental in training a specialized shallow multi-label neural network. The neural network features a single hidden layer with 50 neurons. This neural network makes the final class prediction. Furthermore, the framework includes a strategy for handling records from unknown networks, employing a threshold of 0.8 to determine the network's confidence level. If this threshold is not met, the sample is labeled with $\bm{N}$, indicating an unknown grid origin.

\section{\uppercase{Experimental Evaluation}}
\label{sec:results}

The experimental evaluation of the fusion framework\footnote{\url{https://github.com/GeorgeJoLo/ENFusion}} is detailed, encompassing the description of the training and testing phases. The framework's performance is also assessed against state-of-the-art methods, employing the 2016 SP Cup dataset.

\begin{table*}[!ht]
\centering
\caption{Performance comparison of various classification methods on the 2016 SP Cup dataset.}
\setlength{\tabcolsep}{45pt}
\begin{tabular}{|c|c|}
  \hline
  \textbf{Method} & \textbf{Accuracy} \\
  \hline
  \multirow{2}{*}{SVM, One-vs-One} & \multirow{2}{*}{86\%} \\ 
  & \\ \cite{triantafyllopoulos2016exploring} & \\
  \hline
  \multirow{2}{*}{Multiclass SVM} & \multirow{2}{*}{77\%} \\
  & \\ \cite{ohib2017enf} & \\
  \hline
  \multirow{2}{*}{Random Forrest, SVM, and AdaBoost} & \multirow{2}{*}{88\%} \\
  & \\ \cite{el2016anovel} & \\
  \hline
  \multirow{2}{*}{Binary SVM} & \multirow{2}{*}{87\%} \\
  & \\ \cite{despotovic2016exploring} & \\
  \hline
  \multirow{2}{*}{Multi-Harmonic Histogram Comparison} & \multirow{2}{*}{88\%} \\
  & \\ \cite{chow2016multi-harmonic} & \\
  \hline
  \multirow{2}{*}{Multiclass SVM} & \multirow{2}{*}{88\%} \\
  & \\ \cite{zhou2016geographic} & \\
  \hline
  \multirow{2}{*}{\textbf{Fusion Framework (Here)}} & \multirow{2}{*}{\textbf{96}\%} \\
  & \\
  \hline 
\end{tabular}
\label{tab:discussion}
\end{table*}

\subsection{Model Training and Testing}\label{subsec:training_testing}

The training process initiates with the partitioning of the training dataset, allocating 80\% for training the five individual classifiers, as described in Section~\ref{subsec:classifiers}, and reserving the remaining for training the fusion model, elaborated in Section~\ref{subsec:fusion}. Within these datasets, 20\% is set aside for model validation. Each training set is distinct, following the data bagging method in the context of the five classifiers.

For CNN training (see Table~\ref{tab:cnn_arch}) to attain the highest classification performance, a NAS is conducted using the $\texttt{Optuna}$ library \cite{akiba2019optuna}. The search involves adjusting hyperparameters, like the number of dense units, learning rate, and optimizer values. In the optimization process, the learning rate and the parameters for the Adaptive Moment Estimation (Adam) optimizer \cite{kingma2014adam} were subject to fine-tuning. Initially, the learning rate was set within a range from $10^{-4}$ to $10^{-2}$, and the $\beta$ values for the Adam optimizer varied between 0.9 to 0.999 and 0.99 to 0.999, respectively. After optimization with the $\texttt{Optuna}$ library, the ideal settings were established as a learning rate of $7.2 \times 10^{-4}$, with $\beta_1$ at 0.98, influencing the exponential decay rate for the first moment estimates, and $\beta_2$ at 0.99, impacting the second-moment estimates in the Adam optimizer. This configuration helps in balancing the influence of past and current gradients for efficient optimization. Additionally, the effectiveness of the CNN was further enhanced by integrating extra convolutional and dropout layers, significantly improving its performance and generalization capabilities.

The training procedure for each model, including fusion, is iterated 20 times, and the average validation accuracy is summarized in Table~\ref{tab:validation_acc}.
The table provides insights into the performance of these classifiers across both audio and power data classification tasks. 

Across the audio classification task, it is evident that the fusion model achieves the best classification accuracy in 8 out of the 9 classes. However, for the class of the grid $\bm{G}$, the optimized CNN outperforms the proposed fusion framework. On the other hand, when power data are employed, the fusion model demonstrates its prowess by achieving the best classification accuracy in 8 classes. 

In this context, each classifier achieves commendable accuracy individually. Nevertheless, within the fusion model, a comprehensive solution emerges, showcasing a collective synergy that consistently outperforms the performance of the classifiers when employed separately. These results underscore the fusion model's adaptability in addressing a range of classification challenges by leveraging the strengths of the five classifiers.

The efficacy of the proposed fusion framework is assessed based on the accuracy attained across the 100 testing samples, as detailed in Table~\ref{tab:testing_acc}. Notably, the framework accurately predicts all samples for power recordings since the overall accuracy for them is 100\%. Among the forty audio samples, four are misclassified, resulting in a measured accuracy of 90\% for the audio recordings, which inherently pose a greater difficulty in recognition due to the weaker ENF traces. Additionally, except for one error in class $\bm{H}$, associated with the lowest fusion model accuracy rate for audio, misclassifications occur for grids outside the known $\bm{A}$-$\bm{I}$, that should have been classified as $\bm{N}$. In summary, the proposed fusion framework achieves an overall accuracy of 96\% across the entire testing set.

\subsection{Discussion}

In Table~\ref{tab:discussion}, various classification methods are developed, which focus on power grid classification using the 2016 SP Cup data incorporating the state-of-the-art methods that exhibit varying levels of performance in power grid classification. Commonly shared among these methods is the application of statistical analysis techniques, including utilizing statistical moments and incorporating wavelet features and window feature extraction methods during the processing of the extracted ENF signal.

The proposed fusion framework achieves a 96\% accuracy, outperforming the competitors listed in Table~\ref{tab:discussion}. The accuracies presented in Table~\ref{tab:discussion} are derived from the evaluations using the test set as in Table~\ref{tab:testing_acc}. This achievement is not only a testament to the framework's capabilities for data preprocessing analysis but also encompasses steps like audio augmentation and spectrogram generation. Moreover, utilizing the focused spectrogram within the fusion of classifiers adds an extra layer of robustness, indicating the model's competence in addressing the challenges posed by power grid classification.

The proposed fusion framework demonstrates limited proficiency in recognizing records from grids not included in the training dataset. Table~\ref{tab:testing_acc} illustrates this constraint, wherein among the ten samples expected to belong to class $\bm{N}$, three are inaccurately classified into other classes. These samples should perform ENF signals similar to already known girds. This observation highlights a significant challenge in grid identification, underscoring the subtle nature of ENF differences among distinct grids. Additionally, identifying samples originating from unknown grids necessitates a dedicated study, indicating an avenue for further exploration and refinement in future research endeavors.

\section{\uppercase{Conclusions}}
\label{sec:conclusion}

In this paper, a novel fusion framework for power grid classification has been proposed. The fusion framework, which integrates a CNN optimized via NAS with four traditional machine learning classifiers, has significantly advanced this field. The unique strategy of data augmentation and transformation of audio and power samples into spectrograms has been effectively utilized, focusing on the nominal frequencies to enhance the robustness and accuracy of the model. Furthermore, employing a One-vs-All classification strategy has been instrumental in achieving superior accuracy rates in both training and testing phases, outperforming the state-of-the-art methods. This approach has amplified the model's effectiveness in distinguishing between different grids and contributed to its robustness against overfitting. Future research could focus on collecting and integrating data from additional power grids, thereby expanding the dataset and offering a more thorough evaluation of the proposed methodology's efficacy.

\section*{\uppercase{Acknowledgements}}

This research was supported by the Hellenic Foundation for Research and Innovation (H.F.R.I.) under the “2nd Call for H.F.R.I Research Projects to support Faculty Members \& Researchers” (Project Number: 3888).

\bibliographystyle{apalike}
{\small
\bibliography{example}}

\end{document}